\documentclass[letterpaper]{article} % DO NOT CHANGE THIS
\usepackage{aaai25}  % DO NOT CHANGE THIS
\usepackage{times}  % DO NOT CHANGE THIS
\usepackage{helvet}  % DO NOT CHANGE THIS
\usepackage{courier}  % DO NOT CHANGE THIS
\usepackage[hyphens]{url}  % DO NOT CHANGE THIS
\usepackage{graphicx} % DO NOT CHANGE THIS
\usepackage{natbib}  % DO NOT CHANGE THIS AND DO NOT ADD ANY OPTIONS TO IT
\usepackage{caption} % DO NOT CHANGE THIS AND DO NOT ADD ANY OPTIONS TO IT
\usepackage{algorithm}
\usepackage{algorithmic}

\usepackage{multirow}
\usepackage{booktabs}
\usepackage{enumitem}
\usepackage{pifont} % \ding{52}
\usepackage{newfloat}
\usepackage{listings}
\DeclareCaptionStyle{ruled}{labelfont=normalfont,labelsep=colon,strut=off} % DO NOT CHANGE THIS
\floatstyle{ruled}
\newfloat{listing}{tb}{lst}{}
\floatname{listing}{Listing}
\title{Cross-View Referring Multi-Object Tracking}
\author{
    Sijia Chen, En Yu, Wenbing Tao\thanks{Corresponding author}
}
\affiliations{
    Huazhong University of Science and Technology\\
    \{sijiachen, yuen, wenbingtao\}@hust.edu.cn, *: Corresponding author
}

\begin{document}

\twocolumn[{
\renewcommand\twocolumn[1][]{#1}%
\maketitle
\begin{figure}[H]
\centering
\vspace{-11mm}
\hsize=\textwidth
    \includegraphics[width=2.02\columnwidth]{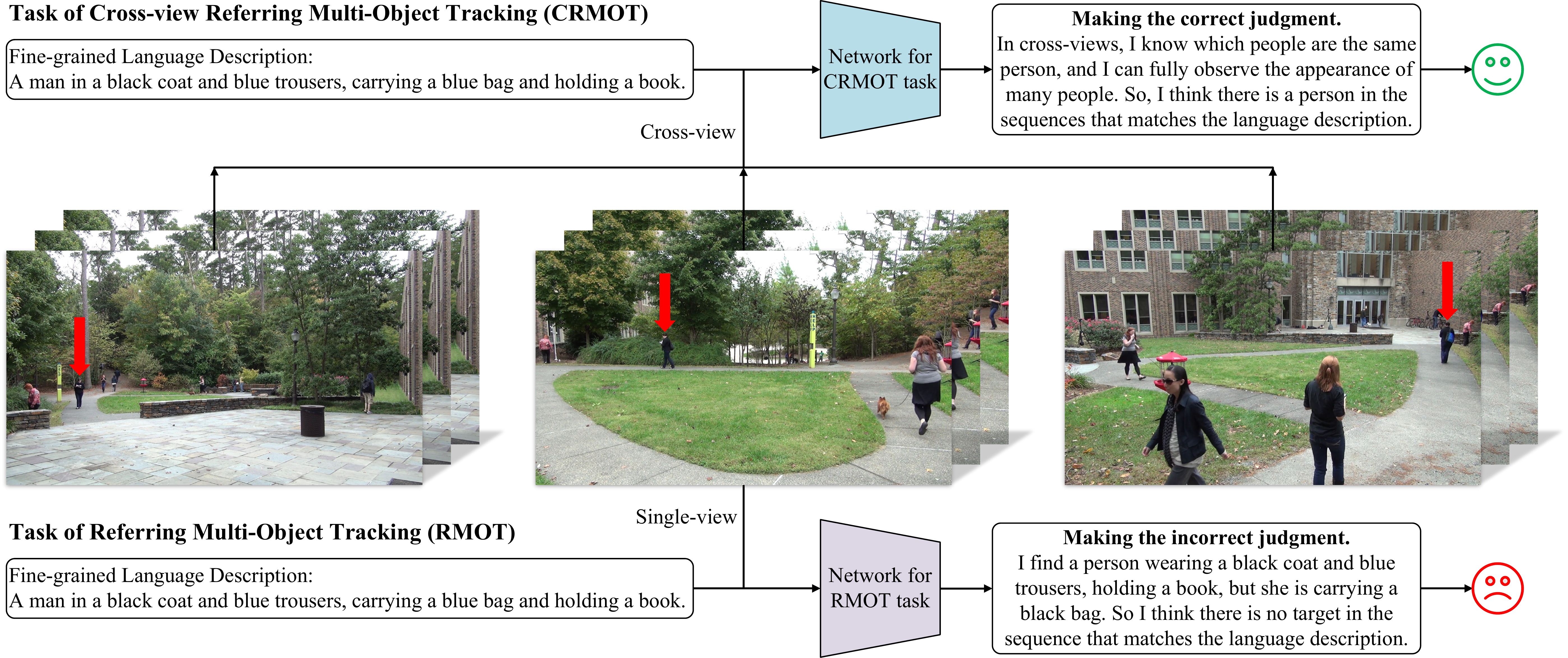}
    % \caption{CRMOT task is the answer to RMOT task.}
    \caption{The difference between CRMOT and RMOT. The CRMOT task introduces the cross-view to obtain the appearances of objects from multiple views, avoiding the problem that the appearances of objects are easily invisible in the RMOT task.
    }
    \label{fig:CRMOT task.}
\end{figure}
}]

\begin{abstract}
Referring Multi-Object Tracking (RMOT) is an important topic in the current tracking field. Its task form is to guide the tracker to track objects that match the language description. Current research mainly focuses on referring multi-object tracking under single-view, which refers to a view sequence or multiple unrelated view sequences. However, in the single-view, some appearances of objects are easily invisible, resulting in incorrect matching of objects with the language description. In this work, we propose a new task, called \textbf{C}ross-view \textbf{R}eferring \textbf{M}ulti-\textbf{O}bject \textbf{T}racking (\textbf{CRMOT}). It introduces the cross-view to obtain the appearances of objects from multiple views, avoiding the problem of the invisible appearances of objects in RMOT task. CRMOT is a more challenging task of accurately tracking the objects that match the language description and maintaining the identity consistency of objects in each cross-view. To advance CRMOT task, we construct a cross-view referring multi-object tracking benchmark based on CAMPUS and DIVOTrack datasets, named \textbf{CRTrack}. Specifically, it provides 13 different scenes and 221 language descriptions. Furthermore, we propose an end-to-end cross-view referring multi-object tracking method, named \textbf{CRTracker}. Extensive experiments on the CRTrack benchmark verify the effectiveness of our method. 
\begin{links}
\link{Dataset, Code}{https://github.com/chen-si-jia/CRMOT}
\end{links}

\end{abstract}

% section1
\section{Introduction}
Multi-Object Tracking (MOT) is one of the most challenging tasks in computer vision. It is widely used in fields such as autonomous driving \cite{li2024fast}, video surveillance \cite{yi2024ucmctrack}, and smart transportation \cite{bashar2022multiple}. Existing MOT methods have already demonstrated effectiveness in addressing most visual generic scenarios. However, when it comes to multimodal contexts, i.e., vision-language scenarios, traditional MOT methods face significant challenges and limitations. To solve this problem, the task of Referring Multi-Object Tracking (RMOT) was recently proposed. The form of this task is to guide the tracker to track the objects that match the language description. For example, if the input “A man in a black coat and blue trousers, carrying a blue bag and holding a book.”, the network for the RMOT task will predict all target trajectories corresponding to that language description. Current research mainly focuses on the RMOT task under the single-view, which refers to a view sequence or multiple unrelated view sequences. However, in the single-view, some appearances of the objects are easily invisible, causing the network for the RMOT task to incorrectly match objects with the fine-grained language description.

To overcome the limitation of the single-view, we propose a new task called Cross-view Referring Multi-Object Tracking (CRMOT). 
It introduces the cross-view, which refers to different views with large overlapping areas, to obtain the appearances of objects from multiple views, thereby avoiding the problem that the appearances of objects are easily invisible in the RMOT task. CRMOT is a more challenging task of accurately tracking the objects that match the fine-grained language description and maintaining the identity (ID) consistency of the objects in each cross-view. 
As illustrated in Figure \ref{fig:CRMOT task.}, we can observe that the network for the RMOT task makes the incorrect judgment when some appearances of the objects are invisible in the single-view of the RMOT task. In contrast, in the cross-views of the CRMOT task, the appearance of the objects can be fully captured, so that the network for the CRMOT task can accurately track the objects that match the fine-grained language description and can know which objects have the same identity (ID) in each cross-view, i.e., the network for the CRMOT task makes the correct judgment.

To advance the research on the cross-view referring multi-object tracking (CRMOT) task, we propose a benchmark, called CRTrack. Specifically, CRTrack includes 13 different scenes, 82K frames, 344 objects, and 221 language descriptions, as detailed in Table \ref{tab:Dataset Statistics of the CRTrack Benchmark.}. These sequence scenes come from two cross-view multi-object datasets, DIVOTrack \cite{hao2024divotrack} and CAMPUS \cite{xu2016multi}. Additionally, we propose a new annotation method based on the unchanging attributes of the objects throughout the sequences. These attributes include headwear color, headwear style, coat color and style, trousers color and style, shoes color and style, held item color, held item style, and transportation. Then, we utilize the large language model GPT-4o to generate language descriptions from the annotated attributes, followed by careful manual checking and correction to ensure the accuracy of language descriptions. Finally, we propose a set of evaluation metrics specifically designed for the CRMOT task.

Moreover, to further advance the research on the CRMOT task, we propose an end-to-end cross-view referring multi-object tracking method, called CRTracker. Specifically, CRTracker combines the accurate multi-object tracking capability of CrossMOT \cite{hao2024divotrack} and the powerful multi-modal capability of APTM \cite{yang2023towards}. Furthermore, a prediction module is designed within the CRTracker network. The novel design idea of this prediction module is to use the frame-to-frame association results of the network as detection results, the fusion scores as confidences, and the prediction module plays the role of a tracker.

Finally, we evaluate our proposed CRTracker method and other methods on the in-domain and cross-domain test sets of the CRTrack benchmark. The evaluation results demonstrate that our method achieves state-of-the-art performance while showing significant generalization capabilities. Specifically, compared to the best-performing method among other single-view approaches, our method surpasses it by 31.45\% in CVRIDF1 and 25.83\% in CVRMA across all scenes in the in-domain evaluation, and by 8.74\% in CVRIDF1 and 1.92\% in CVRMA across all scenes in the cross-domain evaluation.

In summary, our main contributions are as follows:
\begin{itemize} [leftmargin=7.5mm]
    \item[1.] We propose a new task, called Cross-view Referring Multi-Object Tracking (CRMOT). It is a challenging task of accurately tracking the objects that match the language description and maintaining the identity consistency of the objects in each cross-view.
    \item[2.] We construct a benchmark, called CRTrack, to advance the research on the CRMOT task. This benchmark includes 13 different scenes, 82K frames, 344 objects, and 221 language descriptions.
    \item[3.] We propose an end-to-end cross-view referring multi-object tracking method, called CRTracker. We evaluate CRTracker and other methods on the CRTrack benchmark both in-domain and cross-domain. The evaluation results show that CRTracker achieves state-of-the-art performance, fully demonstrating its effectiveness.
\end{itemize}

\begin{figure*}[t]
\centering
\includegraphics[width=1.0\linewidth]{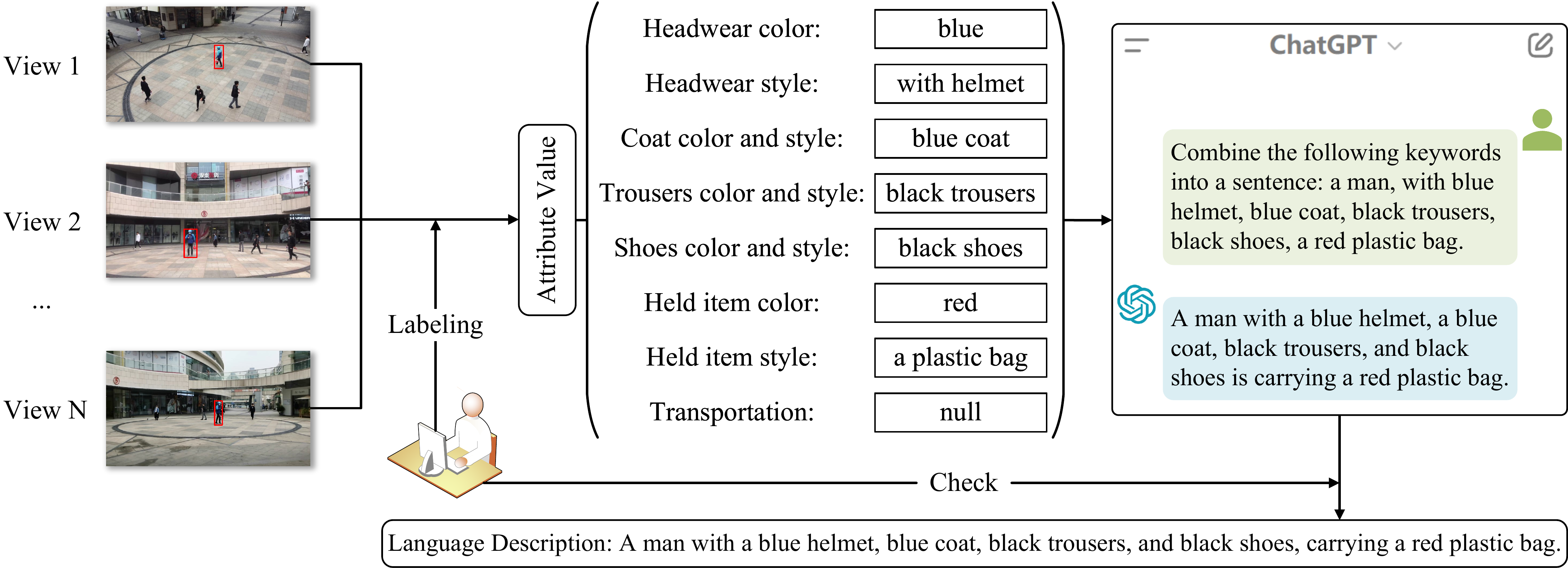} 
\caption{Language Description Annotation Pipeline.}
\label{fig:Language Description Annotation Pipeline.}
\vspace{-2mm}
\end{figure*}

% section2
\section{Related Work}

\noindent \textbf{Cross-View Multi-Object Tracking.}
Cross-view multi-object tracking is a specific category of multi-object tracking \cite{zhang2021fairmot, zeng2022motr, yu2022relationtrack, yu2023generalizing, yu2023motrv3, chen2024delving, gao2024multiple, li2024matching} that shares large overlapping areas between different views. Currently, mainstream methods \cite{cheng2023rest, hao2024divotrack} use appearance and motion features to measure the similarity of the same pedestrians across different views and associate them. There are several commonly used cross-view multi-object tracking datasets, including DIVOTrack \cite{hao2024divotrack}, CAMPUS \cite{xu2016multi}, EPFL \cite{fleuret2007multicamera}, WILDTRACK \cite{chavdarova2018wildtrack}, and MvMHAT \cite{gan2021self}. The DIVOTrack dataset is the latest cross-view multi-object tracking dataset with 10 scenes captured by 3 moving cameras. The CAMPUS dataset contains real scenes captured by static cameras from 3 or 4 different views. The EPFL dataset is one of the traditional cross-view tracking datasets, but its very low resolution makes it difficult to learn the appearance embeddings of objects. The WILDTRACK was shot in a square, but the pedestrian annotations are incomplete. The MvMHAT was shot on a rooftop, but all videos in the dataset use the same scene and the same person. Therefore, we choose the DIVOTrack and CAMPUS datasets to construct the benchmark.

\noindent \textbf{Referring Multi-Object Tracking.}
Referring multi-object tracking is divided into two architectures: two-stage methods and end-to-end methods. The two-stage methods first explicitly extract object trajectories and then select object trajectories that match the language descriptions. The mainstream two-stage methods include iKUN\cite{du2024ikun} and LaMOT \cite{li2024lamot}. The end-to-end methods directly obtain object trajectories that match the language descriptions. The mainstream end-to-end methods include TransRMOT \cite{wu2023referring} and TempRMOT \cite{zhang2024bootstrapping}.

% section3
\section{Benchmark}
To advance the research on the cross-view referring multi-object tracking (CRMOT) task, we construct a cross-view referring multi-object tracking benchmark, named CRTrack. 
Below, we provide details about the CRTrack benchmark.

\noindent\textbf{Dataset Collection.} The emphasized properties of the cross-view referring multi-object tracking dataset are two major elements: \textit{cross-view} and \textit{referring}. \textit{Cross-view} refers to the overlapping area between different camera views, and \textit{referring} refers to the language description. Therefore, based on the cross-view multi-object tracking datasets DIVOTrack \cite{hao2024divotrack} and CAMPUS \cite{xu2016multi}, we add language descriptions to construct the cross-view referring multi-object tracking benchmark, named CRTrack. The DIVOTrack dataset contains data from 10 different real-world scenes, and it is currently the most scene-rich cross-view multi-object tracking dataset. All sequences are captured using three moving cameras and manually synchronized. The CAMPUS dataset contains 3 different scenes with frequent object occlusion problems. All sequences are captured using 3 or 4 static cameras and manually synchronized. It should be noted that we only use their training data, and unify the image sizes and annotation formats of the DIVOTrack and CAMPUS datasets.
% It should be noted that we only use their training set data and resize the images to 1920*1080.

\begin{figure}[t]
\centering
\includegraphics[width=1.0\linewidth]{figures/word_cloud.pdf} 
\caption{Word Cloud.}
\label{fig:Word Cloud.}
\vspace{-1mm}
\end{figure}

\begin{figure*}[t]
\centering
\includegraphics[width=1.0\linewidth]{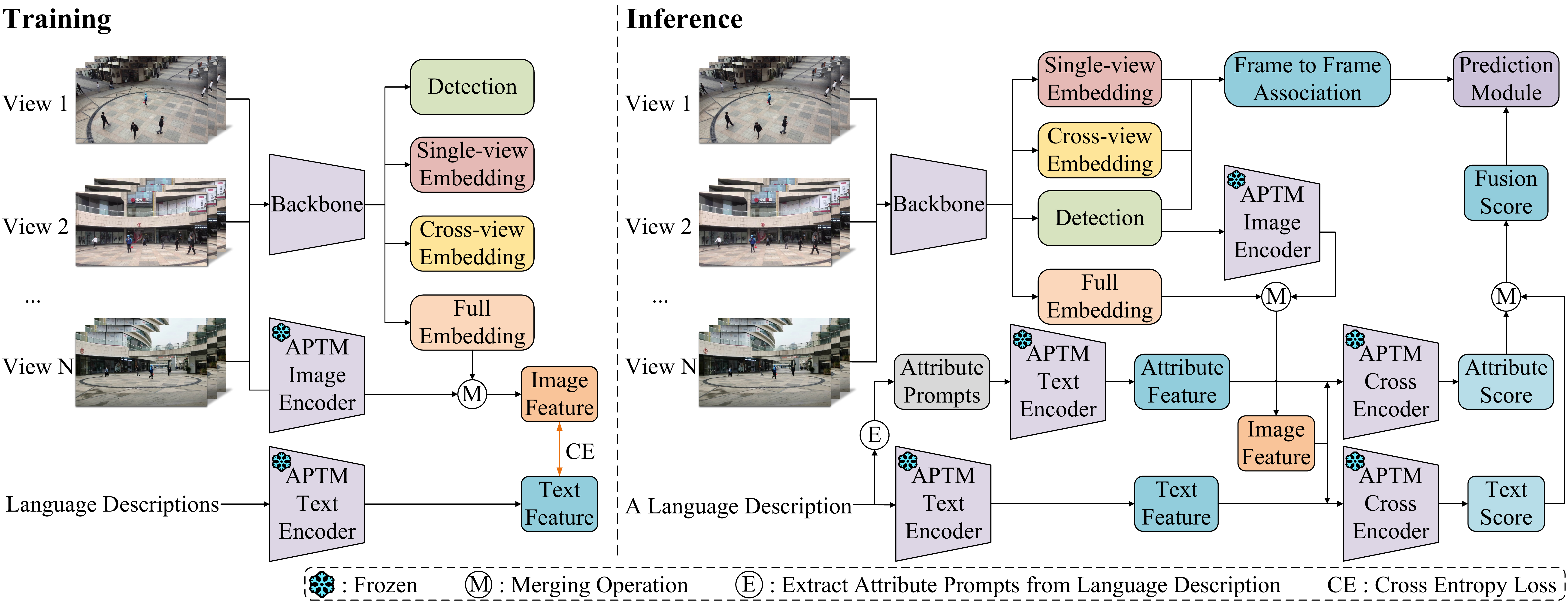}
\caption{Pipeline of CRTracker. It includes a detection head, a single-view Re-ID head, a cross-view Re-ID head, a full Re-ID head and APTM framework. The prediction module outputs the trajectories of objects that match the language description.}
\label{fig:Pipeline of CRTracker.}
\vspace{-1.3mm}
\end{figure*}

\begin{table}[t]
\resizebox{1.0 \columnwidth}{!}{
    \centering
    \begin{tabular}{ccccc}
        \toprule
        \multirow{2}*{Scene} & \multirow{2}*{Views} & Number of & \multirow{2}*{Object Density} & Average number of frames \\
         &  & frames per view &  & of language descriptions \\
        \midrule
        Floor & 3 & 825 & 10.8 & 712 \\
        Gate1 & 3 & 1251 & 9.9 & 780 \\
        Ground & 3 & 901 & 18.9 & 743 \\
        Moving & 3 & 581 & 4.6 & 418 \\
        Park & 3 & 601 & 7.7 & 599 \\
        Shop & 3 & 1101 & 12.5 & 561 \\
        Square & 3 & 601 & 9.1 & 457 \\
        \midrule
        Circle & 3 & 1601 & 8.3 & 952 \\
        Gate2 & 3 & 801 & 3.6 & 685 \\
        Side & 3 & 751 & 12.5 & 624 \\
        \midrule
        Garden1 & 4 & 2849 & 9.6 & 2742 \\
        Garden2 & 3 & 6000 & 5.2 & 3265 \\
        ParkingLot & 4 & 6475 & 4.0 & 3419 \\
        \bottomrule
    \end{tabular}
}
\caption{Dataset Statistics of the CRTrack Benchmark.}
\label{tab:Dataset Statistics of the CRTrack Benchmark.}
\vspace{-1mm}
\end{table}

\noindent \textbf{Dataset Annotation.} We divide the content of the language description into different attributes. These attributes include headwear color, headwear style, coat color and style, trousers color and style, shoes color and style, held item color, held item style, and transportation. Detailed attributes can be found in the \textit{supplementary materials}. Previously, some language descriptions of the RMOT task's benchmark Refer-KITTI \cite{wu2023referring} only annotate a certain fragment sequence of the object, not the whole sequence from the appearance to the disappearance of the object. This annotation method is obviously not suitable for the new task of cross-view referring multi-object tracking, because the introduction of cross-view can observe the whole sequence from the appearance to the disappearance of the object in more detail from multiple views. Therefore, we propose a new annotation method that aims to annotate objects from the perspective of their invariant attributes in the sequence, such as clothing, held items and transportation. We annotate the attributes of the objects in each scene. After obtaining the object annotation attributes, we use the large language model GPT-4o \cite{openai2024gpt4o} to produce the language descriptions based on the object annotation attributes. The language descriptions generated by GPT-4o are manually checked and corrected. With the help of the large language model, the richness of the language descriptions has been greatly improved. Finally, 344 labeled objects and 221 language descriptions are obtained. The entire annotation process is shown in Figure \ref{fig:Language Description Annotation Pipeline.}.

\noindent \textbf{Dataset Split.}
For the DIVOTrack dataset with language descriptions, we evenly selected three scenes as the in-domain test set based on the scene's object density, and the remaining seven scenes as the training set. The CAMPUS dataset with language descriptions is used as the cross-domain test set. In short, the CRTrack benchmark is divided into training set, in-domain test set and cross-domain test set. Specifically, the training set contains "Floor", "Gate1", "Ground", "Moving", "Park", "Shop" and "Square" scenes, the in-domain test set contains "Circle", "Gate2" and "Side" scenes, and the cross-domain test set contains "Garden1", "Garden2" and "ParkingLot" scenes.

\noindent \textbf{Dataset Statistics.}
i) \textbf{Word Cloud.} 
Figure \ref{fig:Word Cloud.} shows the word cloud of the CRTrack benchmark we constructed. We can observe that the CRTrack benchmark contains a large number of words describing clothing, held items and transportation information. The rich variety of word clouds shows the difficulty of our benchmark.
ii) \textbf{Object Density.} 
Object density indicates how many objects there are per frame per cross-view of a scene on average. The object density of each scene in the CRTrack benchmark is shown in Table \ref{tab:Dataset Statistics of the CRTrack Benchmark.}. We can observe that the CRTrack benchmark has scenes with different object densities.
iii) \textbf{Average Number of Frames of Language Description.}
It indicates the average number of frames in which the object corresponding to each language description appears. Table \ref{tab:Dataset Statistics of the CRTrack Benchmark.} shows the average number of frames of the language description of each scene. The average number of frames of the language description of "ParkingLot" scene of the CRTrack benchmark reaches an astonishing 3419. The extremely long number of frames brings great challenges to the cross-view referring multi-object tracking in the temporal dimension.

\subsection{Evaluation Metrics}
The cross-view tracker is different from the single-view tracker. The cross-view tracker processes multiple views in each batch of synchronized video sequences. The same object should have the same identity (ID) in different views. The standard cross-view multi-object tracking evaluation metrics include the cross-view IDF1 (CVIDF1) and the cross-view matching accuracy (CVMA) \cite{gan2021self}. The definitions of CVIDF1 and CVMA are as follows:
\begin{equation}
CVIDF1 = \frac{2  CVIDP \times CVIDR}{CVIDP + CVIDR},
\end{equation}
\begin{equation}
CVMA = 1 - \left(\frac{\sum_t m_t + fp_t+2 mme_t}{\sum_t gt_t}\right)
\end{equation}
where CVIDP and CVIDR denote the cross-view object matching precision and recall, respectively. $m_t$, $fp_t$, $mme_t$, and $gt_t$ are the numbers of misses, false positives, mismatched pairs, and the total number of objects in all views at time $t$, respectively.

It should be noted that cross-view referring multi-object tracking is different from cross-view multi-object tracking. When predicting non-referring but visible objects, they are considered false positives in our evaluation. When the tracking corresponding to the language description is not good, there will be a lot of false detections. This will make CVMA become a relatively large negative number, resulting in a huge impact on the evaluation metrics. We take a maximum value between CVMA value and 0 to prevent the influence of negative numbers.

We aim to comprehensively evaluate each language description, so we propose new evaluation metrics CVRIDF1 and CVRMA for the cross-view referring multi-object tracking (CRMOT) task, and their value range is 0 to 1. The definitions of the evaluation metrics CVRIDF1 and CVRMA for the CRMOT task we proposed are as follows:
\begin{equation}
CVRIDF1 = \frac{\sum_l CVIDF1}{n_l}
\end{equation}
\begin{equation}
CVRMA = \frac{\sum_l \max (CVMA, 0)}{n_l}
\end{equation}
where $l$ represents a language description and $n_l$ denotes the number of language descriptions.

% section4
\section{Strong Baseline of CRMOT}
The challenge of the CRMOT task is to simultaneously detect and track the objects that match the language description and maintain the identity consistency of the objects in each cross-view. To address the challenge of the CRMOT task, we propose an end-to-end cross-view referring multi-object tracking method, named CRTracker, as a strong baseline.

\subsection{Training}

\noindent \textbf{APTM.}
APTM \cite{yang2023towards} is a framework for joint attribute prompt learning and text matching learning, including image encoder, text encoder and cross encoder. Specifically, the image encoder uses Swin Transformer \cite{liu2021swin} to output image features. The text encoder uses the first 6 layers of BERT \cite{devlin2018bert} to output text features. The cross encoder adopts the last 6 layers of BERT, fuses image features and text features, and captures semantic relationships by the cross-attention mechanism.

\noindent \textbf{Pipeline of Training.}
The pipeline of our training framework is shown in Figure \ref{fig:Pipeline of CRTracker.}. The input is synchronized video sequences from multiple cross-views and language descriptions. Similar to the CrossMOT \cite{hao2024divotrack} algorithm, our model uses CenterNet \cite{zhou2019objects} as the backbone, followed by four heads, including a detection head, a single-view Re-ID head, a cross-view Re-ID head and a full Re-ID head. In addition, it also includes APTM image encoder and APTM text encoder. It is worth noting that for single-view Re-ID, the same object in different views is considered as different objects in single-view tracking; for cross-view Re-ID, the same object in different views is considered as the same object; the full Re-ID head is used for language description calculation. We use the image encoder of APTM to encode the object ground truth area in the input video sequence into the feature ${F}_{Ai}$. Then, the feature ${F}_{Ai}$ is merged with the feature ${F}_{f}$ output by the full Re-ID head to obtain the object image feature ${F}_{i}$. Mathematically, the merging operation can be formulated as follows:
\begin{equation}
{F}_{i} = {F}_{f} + \alpha {F}_{Ai}
\label{eq:equationFi}
\end{equation}
where $\alpha$ represents the feature fusion weight of ${F}_{Ai}$.

\noindent Additionally, we use the text encoder of APTM to encode language descriptions and obtain text features. The object image features and text features are calculated using the referring loss ${L}_{r}$. The detection, single-view Re-ID and cross-view Re-ID are calculated using the loss ${L}_{cmot}$.

\noindent \textbf{Loss Functions.}
Our cross-view referring multi-object tracking loss ${L}_{crmot}$ is divided into two parts, cross-view multi-object tracking loss ${L}_{cmot}$ and referring loss ${L}_{r}$.

The ${L}_{cmot}$ is formulated as follows:
\begin{equation}
{L}_{cmot} = \frac{1}{2}\left(\frac{1}{e^{w_1}} {L}_d+\frac{1}{e^{w_2}}\left({L}_s+{L}_c\right)+w_1+w_2\right)
\end{equation}
where ${L}_d$ represents the detection loss, ${L}_s$ represents the single-view Re-ID loss, ${L}_c$ represents the cross-view Re-ID loss. $w_1$ and $w_2$ are are learnable parameters.

The ${L}_{r}$ uses the Cross-Entropy Loss \cite{zhang2018generalized}, which is formulated as:
\begin{equation}
{L}_{r} = -\frac{1}{N} \sum_{i=1}^N \sum_{j=1}^K y_{i,j} \log \left(p_{i,j}\right)
\end{equation}
where $N$ represents the number of objects, $K$ represents the number of all language descriptions in the training data, $y_{i,j}$ represents the label of the $j$-th language description corresponding to the $i$-th object, $p_{i,j}$ represents the probability that the $i$-th object is predicted to be the $j$-th label value.

Thus, the final loss ${L}_{crmot}$ is:
\begin{equation}
{L}_{crmot} = {L}_{cmot} + {L}_{r}
\end{equation}
where ${L}_{crmot}$ represents the cross-view referring multi-object tracking loss.

\begin{algorithm}[t] % tb
\caption{Prediction Module}
\label{alg:Prediction Module}
\textbf{Input}: Frame-to-frame association results, i.e. input tracks of the prediction module $\mathcal{T}_{input}$; fusion scores $S_{f}$\\
\textbf{Parameter}: Fusion scores of views where the track exists $\mathcal{S}$; fusion score of the track ${S}_{f}$; $j$-th view ${V}_{j}$; Number of views for the track ${N}_{V}$; threshold of average fusion score ${T}_{as}$; threshold of single-view fusion score ${T}_{ss}$; threshold of hit score ${T}_{hs}$; hit score of the track ${S}_{\mathcal{T}_{i}}^{H}$; average hit score ${s}_{1}$; single-view hit score ${s}_{2}$; single-view miss score ${s}_{3}$ \\
\textbf{Output}: Output tracks of the prediction module $\mathcal{T}_{output}$

\begin{algorithmic}[1] %[1] enables line numbers
\STATE Let $\mathcal{T}_{output} \leftarrow \emptyset$; $\mathcal{S} \leftarrow \emptyset$.
\FOR{$\mathcal{T}_{i} \in \mathcal{T}_{input}$}
    % \STATE /* summarize fusion score and number of views of the track $\mathcal{T}_{i}$ in each view */
    \STATE /* summarize scores and view number of the track */
    \STATE ${N}_{V}=0$
    \FOR{${V}_{j} \in \left \{{V}_{1},...,{V}_{N}\right \}$}
        \IF {$\mathcal{T}_{i}$ exists in ${V}_{j}$}
            \STATE $\mathcal{S} \leftarrow \mathcal{S} \cup {S}_{f}$
            \STATE ${N}_{V}+=1$
        \ENDIF
    \ENDFOR
    \STATE /* use scores to filter the track */
    \IF {$(\sum \mathcal{S}) / {N}_{V} > {T}_{as}$}
        \STATE ${S}_{\mathcal{T}_{i}}^{H} += {s}_{1}$
        \STATE $\mathcal{T}_{output} \leftarrow \mathcal{T}_{output} \cup \mathcal{T}_{i}$
    \ELSE
        \FOR{${S}_{f} \in \mathcal{S}$}
            \IF {${S}_{f} > {T}_{ss}$}
                \STATE $\lambda =$ int$({S}_{f} / {T}_{ss})$
                \STATE ${S}_{\mathcal{T}_{i}}^{H} += \lambda {s}_{2}$
            \ELSE
                \STATE ${S}_{\mathcal{T}_{i}}^{H} -= {s}_{3}$
                \STATE ${S}_{\mathcal{T}_{i}}^{H} = \max({S}_{\mathcal{T}_{i}}^{H}, 0)$
            \ENDIF
        \ENDFOR
        \IF {${S}_{\mathcal{T}_{i}}^{H} > {T}_{hs}$}
            \STATE $\mathcal{T}_{output} \leftarrow \mathcal{T}_{output} \cup \mathcal{T}_{i}$
        \ENDIF
    \ENDIF
\ENDFOR
\STATE \textbf{return} $\mathcal{T}_{output}$
\end{algorithmic}
\end{algorithm}

\begin{table*}[ht]
\resizebox{2.12 \columnwidth}{!}{
    \centering
    \begin{tabular}{cc|c|cc|cc|cc|cc}
        \toprule
        \multicolumn{11}{c}{\textbf{In-domain Evaluation}}\\
        \toprule
        \multirow{2}*{Method} & \multirow{2}*{Published} & \multirow{2}*{Epochs} & \multicolumn{2}{c|}{All scenes} & \multicolumn{2}{c|}{Circle} & \multicolumn{2}{c|}{Gate2} & \multicolumn{2}{c}{Side} \\
        \cline{4-11}
        \rule{0pt}{10pt} & & & CVRIDF1↑ & CVRMA↑ & CVRIDF1↑ & CVRMA↑ & CVRIDF1↑ & CVRMA↑ & CVRIDF1↑ & CVRMA↑ \\
        \midrule
        TransRMOT \cite{wu2023referring} & CVPR2023 & 20 & 23.30 & 8.03 & 18.85 & 6.94 & 68.03 & 28.51 & 14.33 & 2.65 \\
        TransRMOT \cite{wu2023referring} & CVPR2023 & 100* & 17.72 & 5.17 & 16.74 & 5.52 & 33.03 & 14.87 & 13.92 & 1.48 \\
        TempRMOT \cite{zhang2024bootstrapping} & arXiv2024 & 20 & 22.18 & 8.62 & 17.53 & 5.66 & 62.08 & 36.00 & 15.09 & 3.43 \\
        TempRMOT \cite{zhang2024bootstrapping} & arXiv2024 & 60* & 23.43 & 10.14 & 20.26 & 8.49 & 63.86 & 45.18 & 14.17 & 0.65 \\
        \textbf{CRTracker(Ours)} & - & 20 & \textbf{54.88} & \textbf{35.97} & \textbf{58.38} & \textbf{42.44} & \textbf{91.60} & \textbf{73.40} & \textbf{37.97} & \textbf{14.87} \\
        \bottomrule
        \toprule
        \multicolumn{11}{c}{\textbf{Cross-domain Evaluation}}\\
        \toprule
        \multirow{2}*{Method} & \multirow{2}*{Published} & \multirow{2}*{Epochs} & \multicolumn{2}{c|}{All scenes} & \multicolumn{2}{c|}{Garden1} & \multicolumn{2}{c|}{Garden2} & \multicolumn{2}{c}{ParkingLot} \\
        \cline{4-11}
        \rule{0pt}{10pt} & & & CVRIDF1↑ & CVRMA↑ & CVRIDF1↑ & CVRMA↑ & CVRIDF1↑ & CVRMA↑ & CVRIDF1↑ & CVRMA↑ \\
        \midrule
        TransRMOT \cite{wu2023referring} & CVPR2023 & 20 & 3.66 & 0.20 & 2.85 & 0.01 & 4.23 & 0.55 & 3.87 & 0 \\
        TransRMOT \cite{wu2023referring} & CVPR2023 & 100* & 2.15 & 0 & 2.22 & 0 & 2.23 & 0 & 1.97 & 0 \\
        TempRMOT \cite{zhang2024bootstrapping} & arXiv2024 & 20 & 3.78 & 0.39 & 3.86 & 0.29 & 2.91 & 0.65 & 4.68 & 0.19 \\
        TempRMOT \cite{zhang2024bootstrapping} & arXiv2024 & 60* & 2.68 & 0.40 & 2.20 & 0 & 2.17 & 0.75 & 3.77 & 0.42 \\
        \textbf{CRTracker(Ours)} & - & 20 & \textbf{12.52} & \textbf{2.32} & \textbf{14.96} & \textbf{2.77} & \textbf{11.87} & \textbf{2.80} & \textbf{10.66} & \textbf{1.30} \\
        \bottomrule
    \end{tabular}
}
\caption{Quantitative results on the in-domain and cross-domain test sets of the CRTrack benchmark. * Indicates that the epoch is the epoch of training in the author's paper. ↑ indicates that higher score is better. The best results are marked in \textbf{bold}.}
\label{tab:Quantitative results.}
\end{table*}

\subsection{Inference}

\noindent \textbf{Pipeline of Inference.}
The pipeline of our inference framework is shown in Figure \ref{fig:Pipeline of CRTracker.}. 
During the inference phase, we process language descriptions one by one. First, multiple cross-view video sequences are input into the network, and the detection head outputs the object bounding boxes. Each bounding box is matched with the corresponding single-view Re-ID features, cross-view Re-ID features, and full Re-ID features. In the frame-to-frame association step, we use the MvMHAT \cite{gan2021self} to associate between frames and multiple cross-views. Next, we use the APTM image encoder to encode the object bounding box areas and obtain the encoded features $F_{Ai}$. The encoded features $F_{Ai}$ is fused with the full Re-ID features $F_{f}$ according to Formula (\ref{eq:equationFi}) to generate the object image features $F_{i}$. Subsequently, the APTM text encoder is used to encode the input language description to obtain the text feature $F_{At}$. Attribute prompts are extracted from the language description and input into the APTM text encoder to obtain the attribute features $F_{Aa}$. Next, the APTM cross encoder is used to process the attribute features $F_{Aa}$ and the image features $F_{i}$, and the attribute scores $S_{a}$ is obtained through the head; the APTM cross encoder is used to process the text feature $F_{At}$ and the image features $F_{i}$, and the text scores $S_{t}$ is obtained through the head. Then, the text scores $S_{t}$ is merged with the attribute scores $S_{a}$ to obtain the fusion scores ${S}_{f}$. Mathematically, the merging operation can be formulated as follows:
\begin{equation}
{S}_{f} = {S}_{t} + \beta e^{{S}_{a}}
\label{eq:equationSf}
\end{equation}
where $\beta$ represents the score fusion weight of $e^{{S}_{a}}$.

\noindent Finally, the frame-to-frame association results and the fusion scores are input into the prediction module to generate the trajectories of objects that match the language description.

\noindent \textbf{Prediction Module.}
The design idea of the prediction module is to regard the frame-to-frame association results as the detection results, the fusion scores $S_{f}$ as the confidences, and the prediction module plays a tracking role. The algorithm of the prediction module is shown in Algorithm~\ref{alg:Prediction Module}.

% section5
\section{Experiments}

\subsection{Settings}
For evaluation, we conduct experiments on the CRTrack benchmark we constructed and follow its evaluation metrics.
Our models are trained for 20 epochs and tested on a single NVIDIA RTX 3090 GPU. The feature dimensions of single-view embedding, cross-view embedding, and full embedding are all set to 512. During the training phase, we use the Adam optimizer \cite{kingma2014adam}, the initial learning rate is set to $1 \times 10^{-4}$, the batchsize to 12, and the feature fusion weight $\alpha$ in Formula (\ref{eq:equationFi}) to 0.01. During the inference phase, we set the score fusion weight $\beta$ in Formula (\ref{eq:equationSf}) is set to 0.1, threshold of average fusion score ${T}_{as}$ to 0.5, threshold of single-view fusion score ${T}_{ss}$ to 0.75, threshold of hit score ${T}_{hs}$ to 30, average hit score ${s}_{1}$ to 3, single-view hit score ${s}_{2}$ to 3, and single-view miss score ${s}_{3}$ to 1.

\begin{figure*}[t]
\centering
\includegraphics[width=1.0\linewidth]{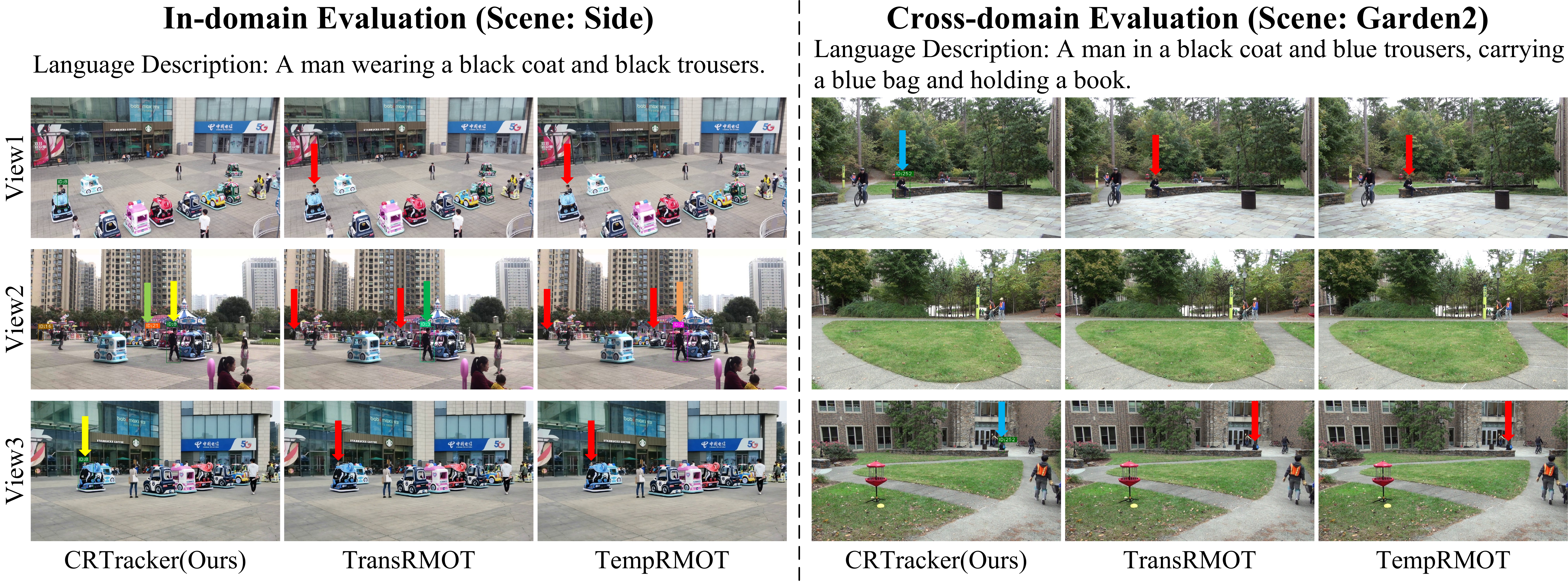} 
\caption{Qualitative results of our proposed CRTracker method and other methods, including TransRMOT and TempRMOT, on the CRTrack benchmark's in-domain and cross-domain evaluations. The rows and columns represent the camera views and different methods, respectively. Red arrows indicate targets that are not correctly detected or matched. Other colored arrows represent correctly detected targets, with arrows of the same color indicating the same target.}
\label{fig:Qualitative Results.}
\end{figure*}
% It should be noted that the objects that match the language description are the targets we need.

\subsection{Quantitative Results}

On the CRTrack benchmark, we compared our CRTracker with other methods. Since previous referring multi-object tracking methods are designed for single-view, they cannot be used for the cross-view referring multi-object tracking task. Thus, we combine previous referring multi-object tracking methods with the MvMHAT \cite{gan2021self} cross-view association algorithm to enable them to be used in the cross-view referring multi-object tracking task. In addition, since our method is end-to-end, for fair comparison, we choose two end-to-end referring multi-object tracking methods, including TransRMOT \cite{wu2023referring} and TempRMOT \cite{zhang2024bootstrapping}. For in-domain evaluation, all methods are trained using the benchmark training set and tested on the benchmark in-domain test set. For cross-domain evaluation, all methods are trained using the benchmark training set and tested on the benchmark cross-domain test set. It is worth noting that our CRTracker and other methods use the same model and parameter settings for cross-domain evaluation as for in-domain evaluation. 

\noindent \textbf{In-domain Evaluation.} As shown in Table \ref{tab:Quantitative results.}, our CRTracker achieves 54.88\% CVRIDF1 and 35.97\% CVRMA on all scenes of the in-domain test set. In particular, it achieves 91.60\% CVRIDF1 and 73.40\% CVRMA on the "Gate2" scene. Notably, our CRTracker far outperforms all other methods in the in-domain evaluation. The results indicate that CRTracker can tackle in-domain scenes well.

\noindent \textbf{Cross-domain Evaluation.} As illustrated in Table \ref{tab:Quantitative results.}, all methods suffer vital performance degradation, which is expected due to the high difficulty of the cross-domain test set of the benchmark. The cross-domain test set and the training set differ in terms of the number of cross views, scenes, pedestrians, camera angles, and lighting. In addition, the cross-domain test set contains many language descriptions that do not appear in the training set, and the average number of frames of language descriptions is very long. Despite this, our CRTracker still surpasses other methods, with achieving 12.52\% CVRIDF1 and 2.32\% CVRMA on all scenes of the cross-domain test set. The results show that CRTracker has a good generalization ability for unseen domains.

\subsection{Qualitative Results}

To further demonstrate the superiority of our CRTracker, we visualize some results of our proposed CRTracker method and other methods trained for 20 epochs in in-domain and cross-domain evaluations. As shown in Figure \ref{fig:Qualitative Results.}, CRTracker is able to accurately detect and track the objects that match the language description in a variety of challenging scenes and keep the same object with the same identity in each cross-view. In the "Garden2" scene example, CRTracker can accurately detect and track the target and keep the target with the same identity in each cross-view even with the untrained language description, which fully demonstrates the generalization capability of our method. Many qualitative results can be found in the \textit{supplementary materials}.

\subsection{Ablation Study}

To study the role of each part of our method CRTracker, we conduct ablation experiments on the CRTrack benchmark. All experiments follow in-domain evaluation, that is, training on the training set and testing on the in-domain test set.

\noindent \textbf{Analysis of Prediction Module.} 
To demonstrate the effectiveness of prediction module, we compare CRTracker with and without prediction module. 
As shown in Table \ref{tab:Results of CRTracker without and with the prediction module.}, we can observe that the CRTracker with prediction module is 7.34\% higher in CVRIDF1 and 7.29\% higher in CVMA than the CRTracker without prediction module. This phenomenon shows that the prediction module fully fuses the trajectory and language description scores from each cross-view to maximize the matching of trajectory to description.

\begin{table}[t]
\centering
\begin{tabular}{c|cc}
    \toprule
    Prediction Module & CVRIDF1↑ & CVRMA↑ \\
    \midrule
    \ding{56} & 47.54 & 28.68 \\
    \ding{52} & \textbf{54.88} & \textbf{35.97} \\
    \bottomrule
\end{tabular}
\caption{Results of CRTracker without and with the prediction module. The best results are marked in \textbf{bold}.}
\label{tab:Results of CRTracker without and with the prediction module.}
\end{table}

% section6
\section*{Acknowledgements}
This work was supported by the National Natural Science Foundation of China under Grants 62176096 and 61991412.

% section7
\section{Conclusion}
In this work, we propose a novel task, named Cross-view Referring Multi-Object Tracking (CRMOT). It is a challenging task of accurately tracking the objects that match the fine-grained language description and maintaining the identity consistency of the objects in each cross-view. To advance the CRMOT task, we construct the CRTrack benchmark. Furthermore, to address the challenge of the new task, we propose CRTracker, an end-to-end cross-view referring multi-object tracking method. We validate CRTracker on the CRTrack benchmark, which achieves state-of-the-art performance and demonstrates good generalization ability.

% Referenes
\bibliography{aaai25}

% Supplementary Material
\section{Supplementary Material}
In this supplementary material, we provide specific details of attributes of our annotation method, as well as more qualitative results in both in-domain and cross-domain evaluations.

\subsection{Details of Attributes}
Our annotation method aims to annotate objects from the perspective of their invariant attributes in the sequence, such as clothing, held items and transportation. We annotate the CRTrack benchmark with 8 attributes and 74 words with specific representations, as shown in Table \ref{tab:Details of attributes.}. The attributes are carefully selected, considering the object's appearance characteristics in the CRTrack benchmark, including headwear color, headwear style, coat color and style, trousers color and style, shoes color and style, held item color, held item style, and transportation. Although the attribute of clothing color may have more than one representation, at most two representations corresponding to the image are selected for annotation. In most cases, the representation that best corresponds to the image is selected for annotation.

\subsection{More Qualitative results}
To further demonstrate the superiority of our proposed CRTracker method, we visualize some results of our method and other methods for each scene in both in-domain and cross-domain evaluations.

Before looking at the visualization results, the following content will help you understand. The rows and columns represent the camera views and different methods, respectively. The red arrows indicate targets that are not correctly detected or matched. Other colored arrows represent correctly detected targets, with arrows of the same color indicating the same target. The double-colored arrows indicate that the IDs of the same target are inconsistent in the time dimension or view dimension. In short, the fewer red arrows, the better the performance.

\begin{figure*}[!p]
\centering
\includegraphics[width=0.95\linewidth]{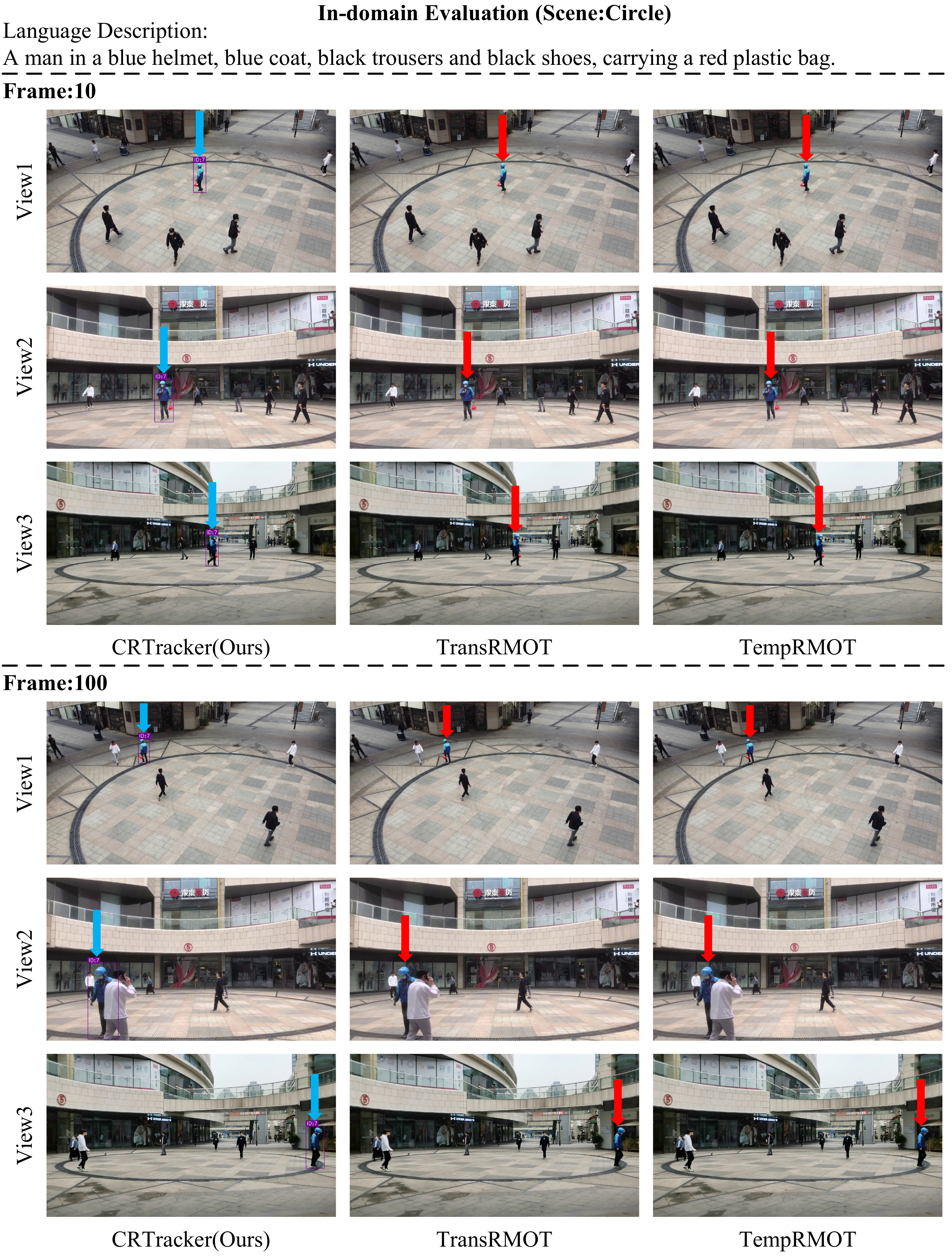} 
\caption{Qualitative results for the “Circle” scene in the in-domain test set. The rows and columns represent the camera views and different methods, respectively. Red arrows indicate targets that are not correctly detected or matched. Other colored arrows represent correctly detected targets, with arrows of the same color indicating the same target. Double-colored arrows indicate that the IDs of the same target are inconsistent in the time dimension or view dimension.}
\label{fig:Circle_2.}
\end{figure*}

\begin{figure*}[!p]
\centering
\includegraphics[width=0.95\linewidth]{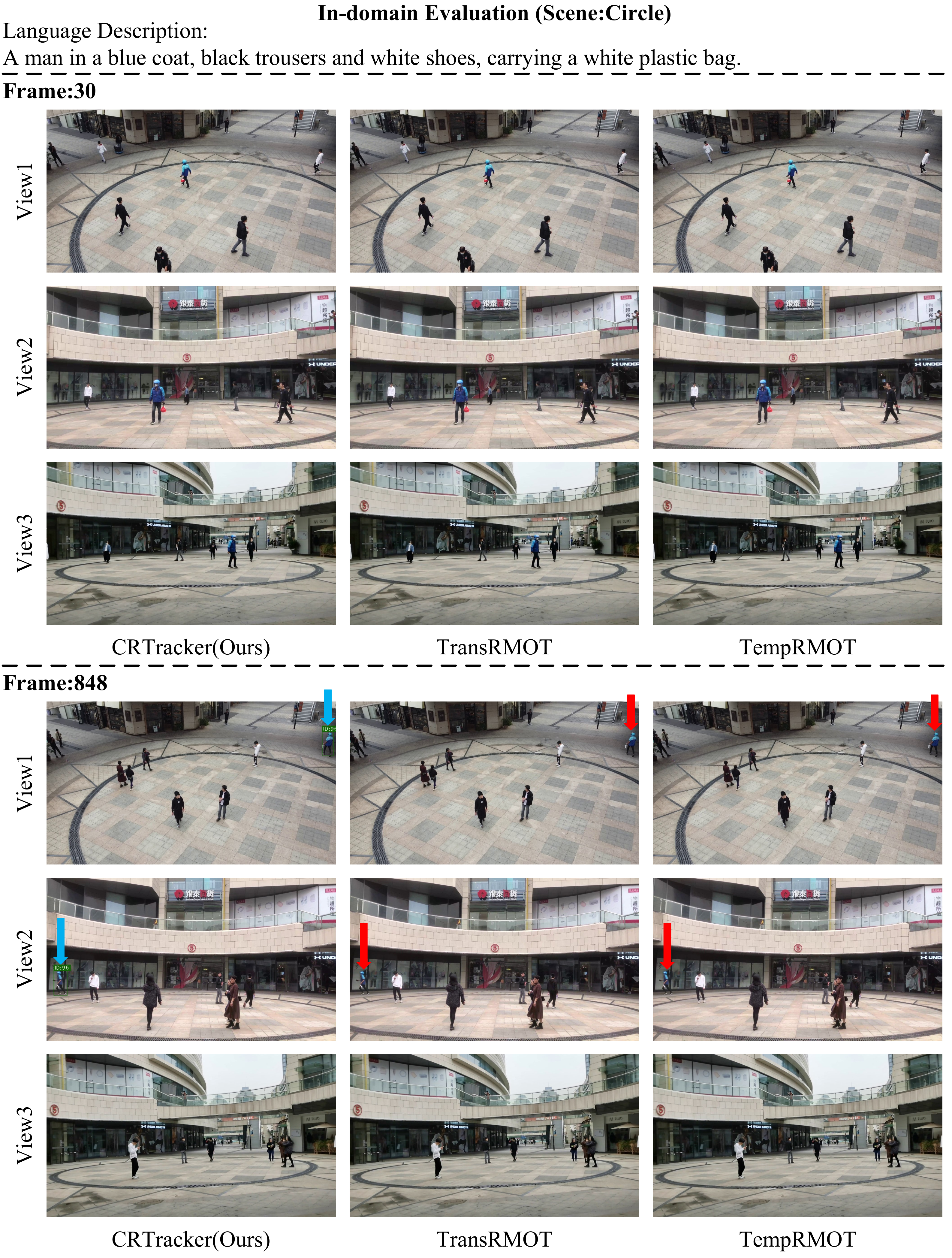} 
\caption{Qualitative results for the “Circle” scene in the in-domain test set. The rows and columns represent the camera views and different methods, respectively. Red arrows indicate targets that are not correctly detected or matched. Other colored arrows represent correctly detected targets, with arrows of the same color indicating the same target. Double-colored arrows indicate that the IDs of the same target are inconsistent in the time dimension or view dimension.}
\label{fig:Circle_3.}
\end{figure*}

\begin{figure*}[!p]
\centering
\includegraphics[width=0.95\linewidth]{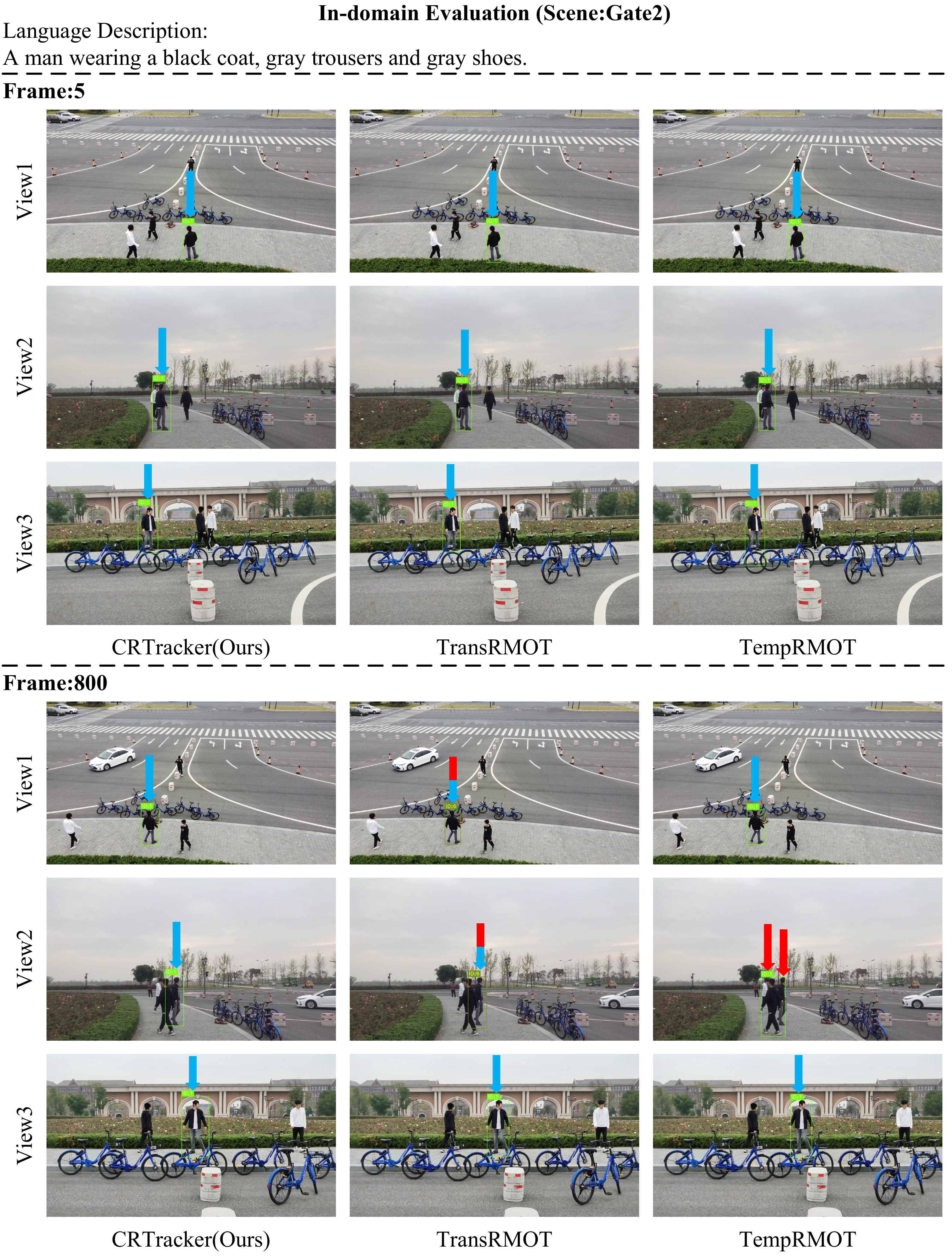} 
\caption{Qualitative results for the “Gate2” scene in the in-domain test set. The rows and columns represent the camera views and different methods, respectively. Red arrows indicate targets that are not correctly detected or matched. Other colored arrows represent correctly detected targets, with arrows of the same color indicating the same target. Double-colored arrows indicate that the IDs of the same target are inconsistent in the time dimension or view dimension.}
\label{fig:Gate2.}
\end{figure*}

\begin{figure*}[!p]
\centering
\includegraphics[width=0.95\linewidth]{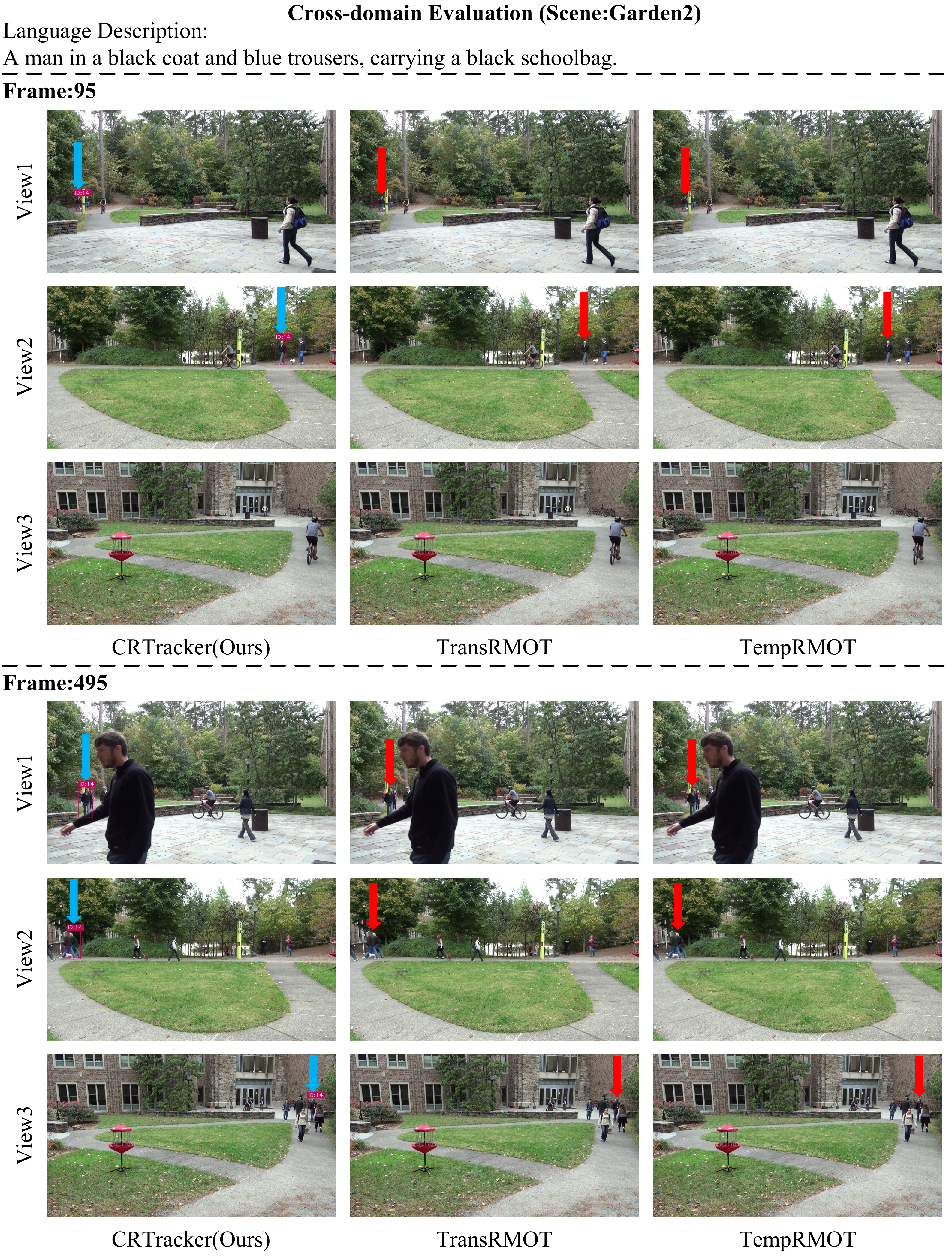} 
\caption{Qualitative results for the “Garden2” scene in the cross-domain test set. The rows and columns represent the camera views and different methods, respectively. Red arrows indicate targets that are not correctly detected or matched. Other colored arrows represent correctly detected targets, with arrows of the same color indicating the same target. Double-colored arrows indicate that the IDs of the same target are inconsistent in the time dimension or view dimension.}
\label{fig:Garden2_1.}
\end{figure*}

\begin{figure*}[!p]
\centering
\includegraphics[width=0.95\linewidth]{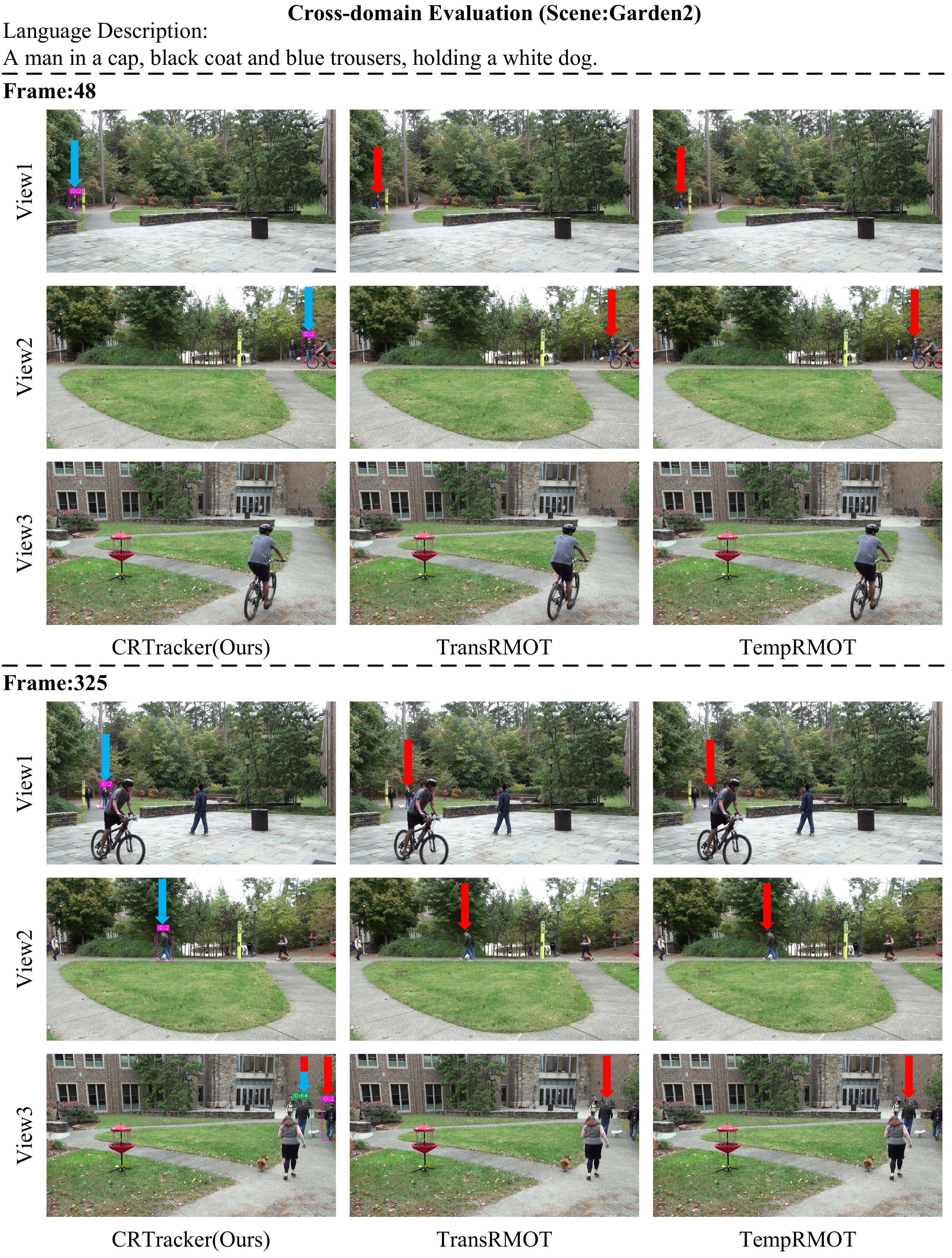} 
\caption{Qualitative results for the “Garden2” scene in the cross-domain test set. The rows and columns represent the camera views and different methods, respectively. Red arrows indicate targets that are not correctly detected or matched. Other colored arrows represent correctly detected targets, with arrows of the same color indicating the same target. Double-colored arrows indicate that the IDs of the same target are inconsistent in the time dimension or view dimension.}
\label{fig:Garden2_2.}
\end{figure*}

\begin{figure*}[!p]
\centering
\includegraphics[width=0.8\linewidth]{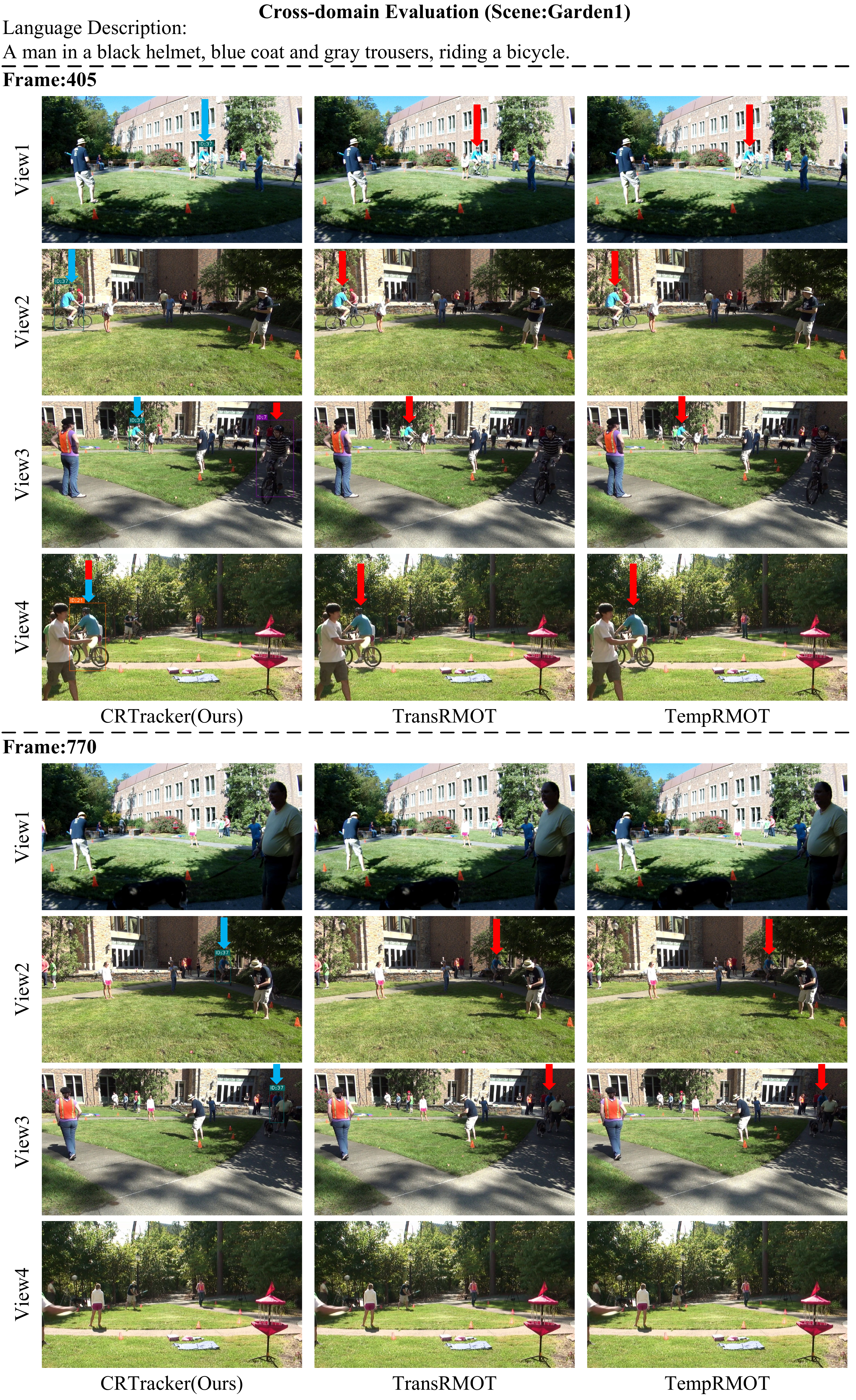} 
\caption{Qualitative results for the “Garden1” scene in the cross-domain test set.}
\label{fig:Garden1_1.}
\end{figure*}

\begin{figure*}[!p]
\centering
\includegraphics[width=0.8\linewidth]{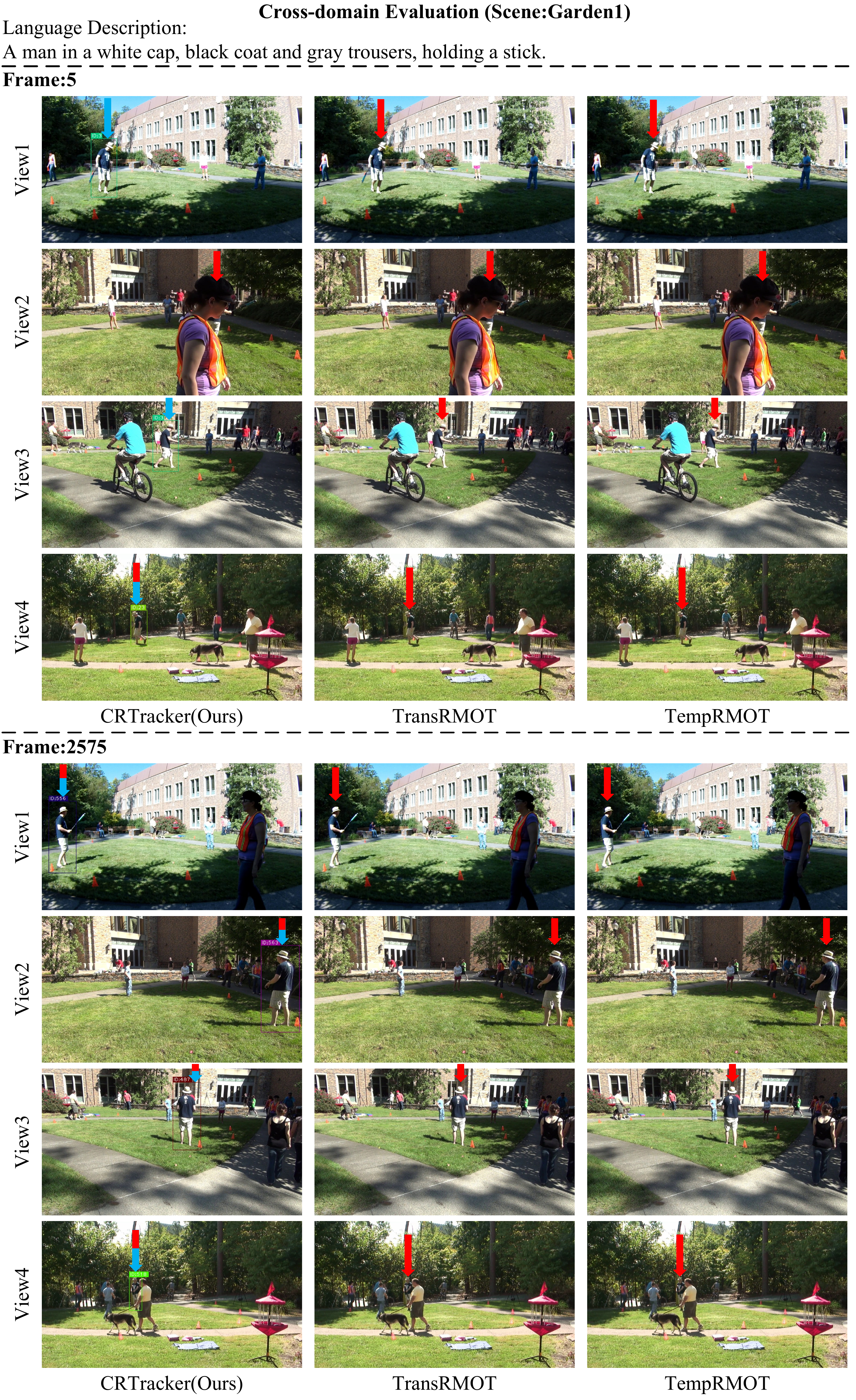} 
\caption{Qualitative results for the “Garden1” scene in the cross-domain test set.}
\label{fig:Garden1_2.}
\end{figure*}

\begin{figure*}[!p]
\centering
\includegraphics[width=0.8\linewidth]{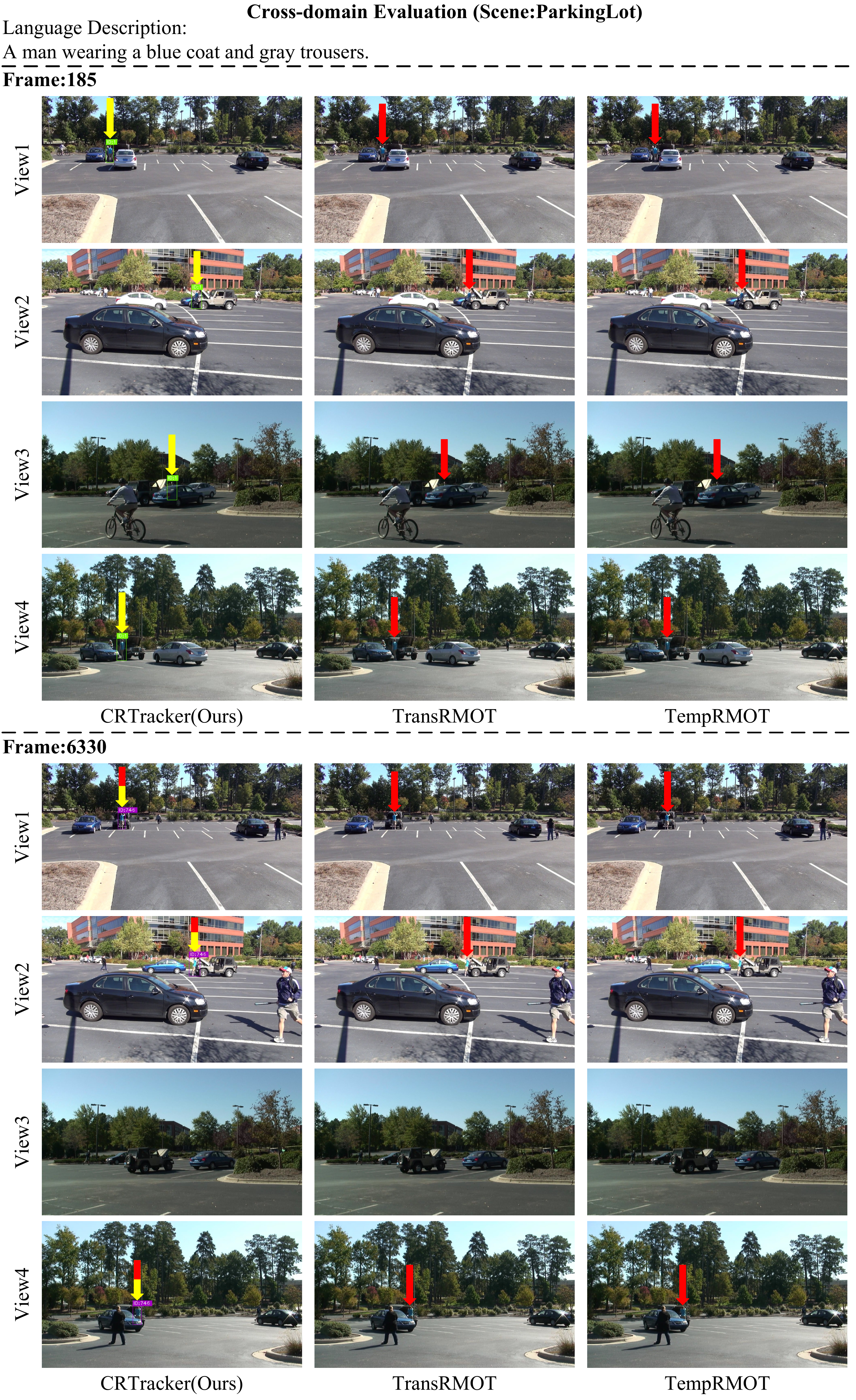} 
\caption{Qualitative results for the “ParkingLot” scene in the cross-domain test set.}
\label{fig:ParkingLot.}
\end{figure*}

Some in-domain evaluation results are shown in Figures \ref{fig:Circle_1.}, \ref{fig:Circle_2.}, \ref{fig:Circle_3.}, \ref{fig:Circle_4.}, \ref{fig:Gate2.} and \ref{fig:Side.}. We can find that our CRTracker shows extremely outstanding performance.
Some cross-domain evaluation results are shown in Figures \ref{fig:Garden2_1.}, \ref{fig:Garden2_2.}, \ref{fig:Garden1_1.}, \ref{fig:Garden1_2.} and \ref{fig:ParkingLot.}. We can observe that other methods cannot find the target at all. 
For the amazing 6330 frames of long tracking in Figure \ref{fig:ParkingLot.}, our method suffers from the ID switching problem.

In summary, Our proposed CRTracker method is able to accurately detect and track the objects that match the fine-grained language description while maintaining the identity (ID) consistency of the objects in each cross-view, providing a good method for the CRMOT task we proposed.

\subsection{Questions and Replies}
\label{sec:Questions and Replies}

\noindent \textbf{(1) We can see that a reference description corresponds to a very long sequence. I wonder whether a reference may describe multiple persons during the same period or different persons at different moments in a long video.}

A reference can describe any number of people at any time who match the language description, as evidenced by Figures \ref{fig:Circle_1.}, \ref{fig:Circle_4.} and \ref{fig:Side.} in the Supplementary Material.

\noindent \textbf{(2) The motivation of the task does not look convincing to me. As long as we can identify the target object from any camera view, why do we have to perform cross-view association?}

For some complex language descriptions, it is difficult to correctly judge whether the target matches the language description from a single view. To overcome the limitation of the single-view, our CRMOT task introduces the cross-view, to obtain the appearances of objects from multiple views, thereby avoiding the problem that the appearances of objects are easily invisible in the single view. For example, in Figure \ref{fig:Qualitative Results.} of the Manuscript, the person indicated by yellow arrows in View1 and View3 is difficult to locate accurately from a single view because some appearances of their appearance are occluded. By introducing cross-views, we can clearly know the same person across all three views and identify the target indicated by the yellow arrow in View2, which allows us to confirm that the persons in View1 and View3 are also the targets we want.

\noindent \textbf{(3) Can CrossMOT be extended as another baseline?}

The CrossMOT method we currently use can be replaced with other networks.

\twocolumn[{
\renewcommand\twocolumn[1][]{#1}%
\begin{table}[H]
\centering
\hsize=\textwidth
\begin{tabular}{l | l}
    \toprule
    \multicolumn{1}{c|}{Attribute} &  \multicolumn{1}{c}{Representation in words} \\
    \midrule
    \rule{0pt}{10pt} Headwear color & "white", "black", "gray", "green", "pink", "red", "yellow", "blue", "orange", "purple", "null"\\
    \hline
    \rule{0pt}{10pt} Headwear style & "with cap", "with helmet", "null"\\
    \hline
    \rule{0pt}{10pt} \multirow{2}*{Coat color and style} & "white coat", "black coat", "gray coat", "green coat", "pink coat", "red coat",\\
        & "yellow coat", "blue coat", "orange coat", "purple coat", "null"\\
    \hline
    \rule{0pt}{10pt} \multirow{2}*{Trousers color and style} & "white trousers", "black trousers", "gray trousers", "green trousers", "pink trousers",\\
        & "red trousers", "yellow trousers", "blue trousers", "orange trousers", "purple trousers", "null"\\
    \hline
    \rule{0pt}{10pt} \multirow{2}*{Shoes color and style} & "white shoes", "black shoes", "gray shoes", "green shoes", "pink shoes", "red shoes",\\
        & "yellow shoes", "blue shoes", "orange shoes", "purple shoes", "null"\\
    \hline
    \rule{0pt}{10pt} Held item color & "white", "black", "gray", "green", "pink", "red", "yellow", "blue", "orange", "purple", "null"\\
    \hline
    \rule{0pt}{10pt} \multirow{2}*{Held item style} & "a bag", "a plastic bag", "a handbag", "a schoolbag", "a cart", "a box", "a child", "a stick",\\
        & "a book", "a mobile phone", "a can", "null"\\
    \hline
    \rule{0pt}{10pt} Transportation & "a bicycle", "an electric bike", "a tricycle", "null"\\
    \bottomrule
\end{tabular}
\caption{Details of attributes.}
\label{tab:Details of attributes.}
\end{table}
}]

% Circle_1
\begin{figure*}[!p]
\centering
\includegraphics[width=0.95\linewidth]{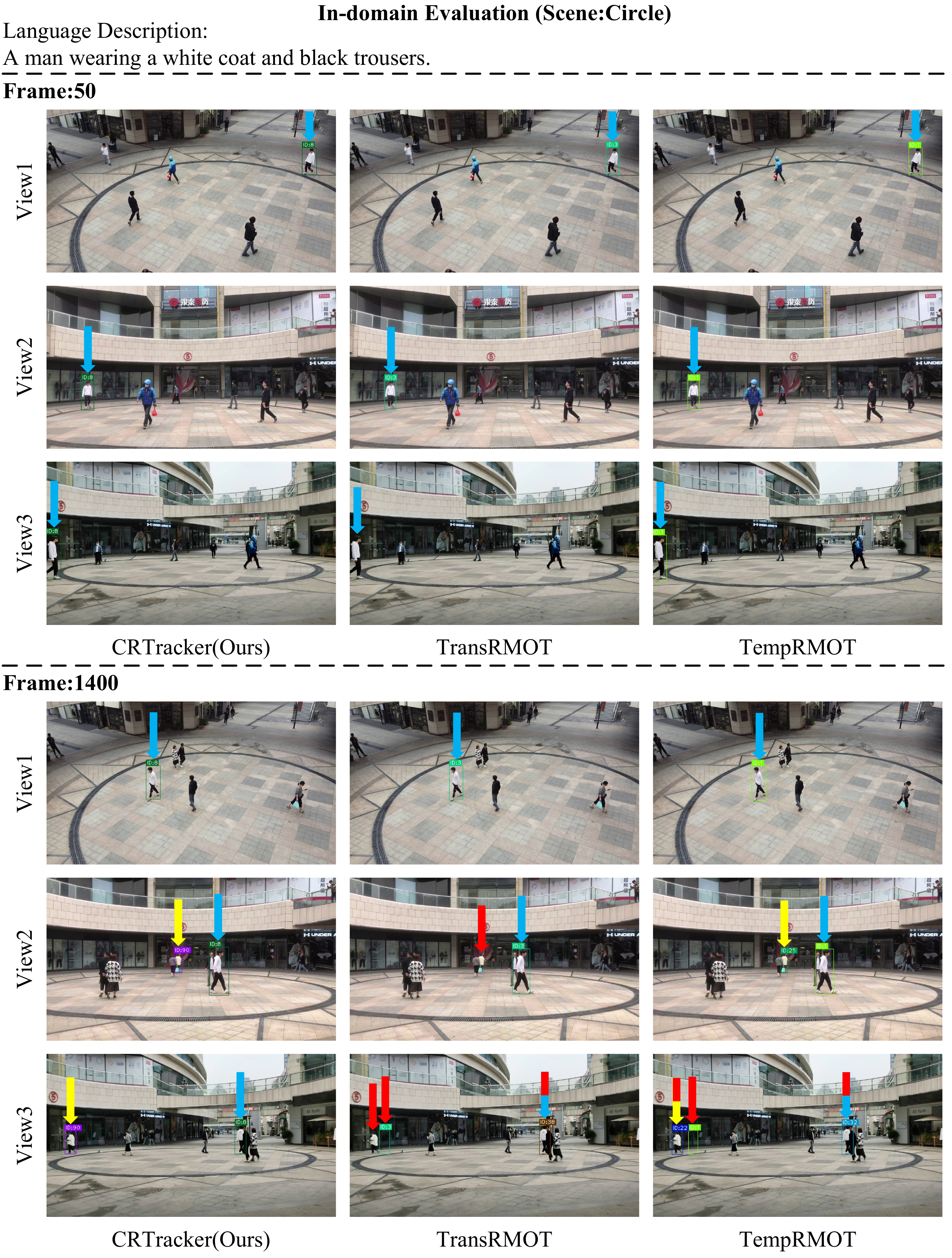} 
\caption{Qualitative results for the “Circle” scene in the in-domain test set. The rows and columns represent the camera views and different methods, respectively. Red arrows indicate targets that are not correctly detected or matched. Other colored arrows represent correctly detected targets, with arrows of the same color indicating the same target. Double-colored arrows indicate that the IDs of the same target are inconsistent in the time dimension or view dimension.}
\label{fig:Circle_1.}
\end{figure*}

% Circle_2

% Circle_3

% Circle_4
\begin{figure*}[!p]
\centering
\includegraphics[width=0.95\linewidth]{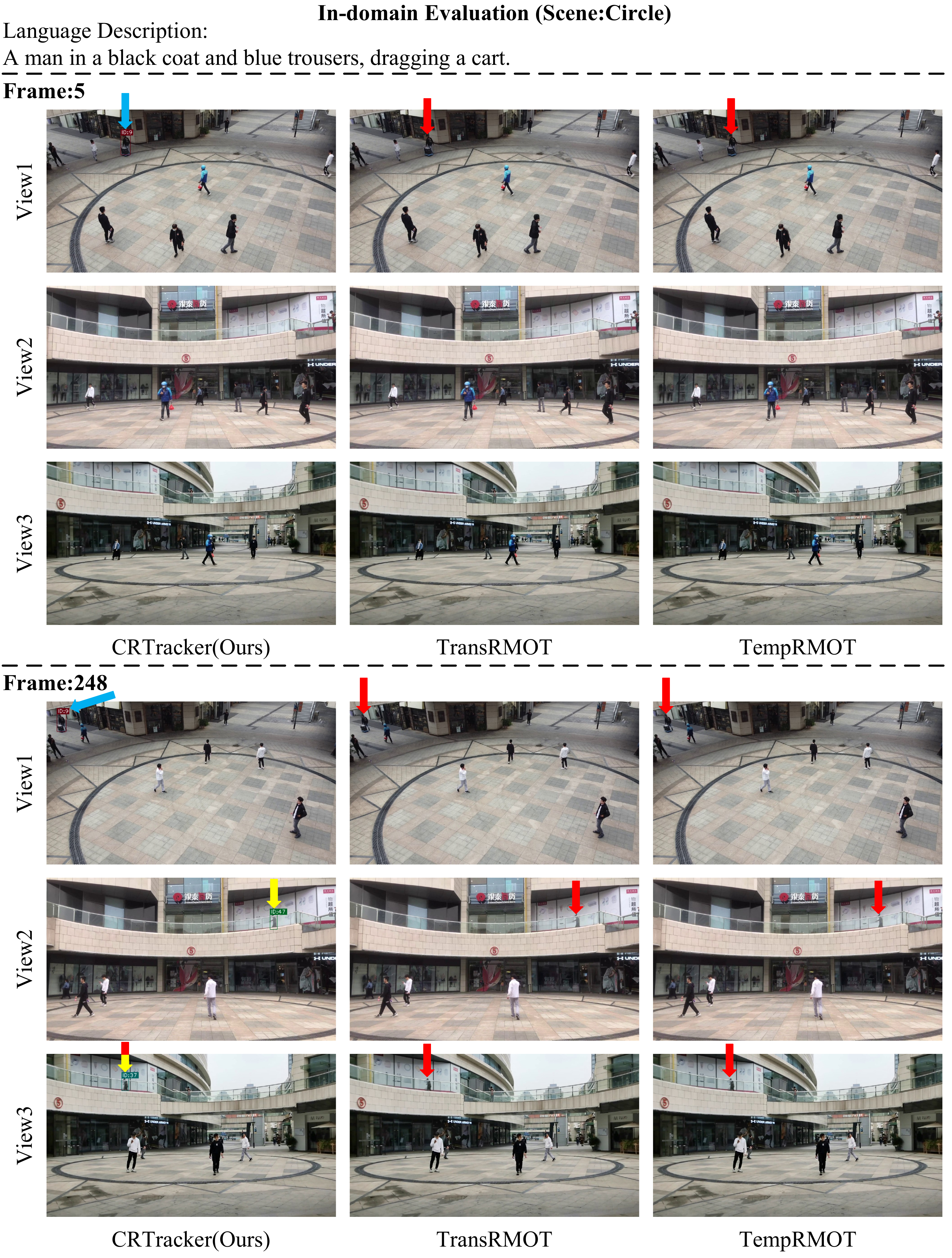} 
\caption{Qualitative results for the “Circle” scene in the in-domain test set. The rows and columns represent the camera views and different methods, respectively. Red arrows indicate targets that are not correctly detected or matched. Other colored arrows represent correctly detected targets, with arrows of the same color indicating the same target. Double-colored arrows indicate that the IDs of the same target are inconsistent in the time dimension or view dimension.}
\label{fig:Circle_4.}
\end{figure*}

% Gate2

% Side
\begin{figure*}[!p]
\centering
\includegraphics[width=0.95\linewidth]{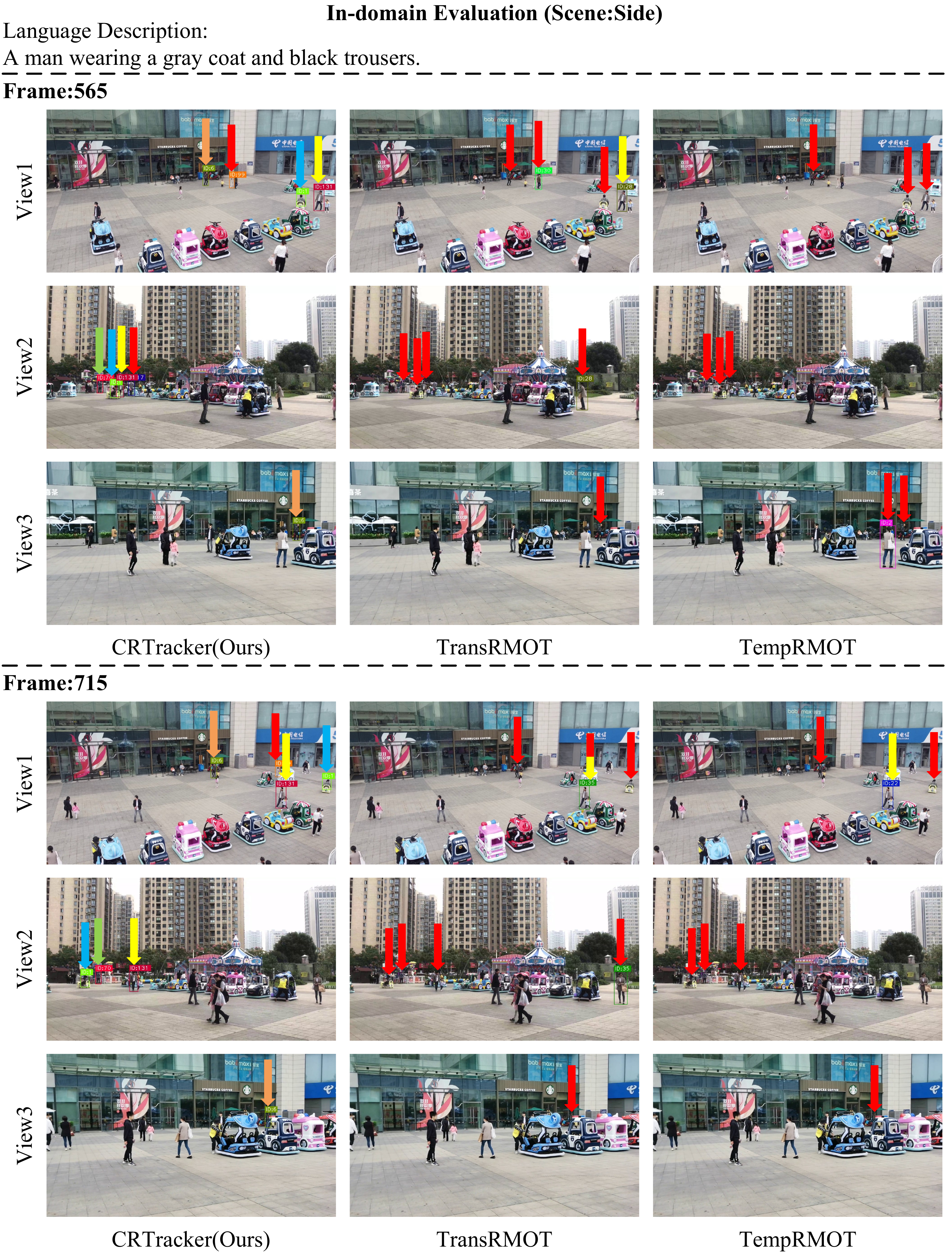} 
\caption{Qualitative results for the “Side” scene in the in-domain test set. The rows and columns represent the camera views and different methods, respectively. Red arrows indicate targets that are not correctly detected or matched. Other colored arrows represent correctly detected targets, with arrows of the same color indicating the same target. Double-colored arrows indicate that the IDs of the same target are inconsistent in the time dimension or view dimension.}
\label{fig:Side.}
\end{figure*}

% Garden2_1

% Garden2_2

% Garden1_1

% Garden1_2

% ParkingLot

\end{document}